\documentclass[11pt]{article}

\usepackage[preprint]{acl}
\usepackage[table]{xcolor}
\usepackage{listings}
\usepackage{tabularx} 
\usepackage{pifont}
\usepackage{booktabs}
\usepackage{xcolor}
\usepackage{enumitem}
\usepackage[most]{tcolorbox}

\definecolor{phgreen}{HTML}{1F77B4}
\definecolor{phorange}{HTML}{FF7F0E}

\usepackage{times}
\usepackage{latexsym}
\usepackage{booktabs}
\usepackage{multirow}
\usepackage[T1]{fontenc}

\usepackage[utf8]{inputenc}

\usepackage{microtype}

\usepackage{inconsolata}

\usepackage{graphicx}
\usepackage{algorithm}
\usepackage{algpseudocode}
\usepackage{amsmath}
\usepackage{amssymb}
\renewcommand{\thefootnote}{\fnsymbol{footnote}}

\title{ArcAligner: Adaptive Recursive Aligner for Compressed Context Embeddings in RAG}

\author{
 \textbf{Jianbo Li\footnotemark[1]},
 \textbf{Yi Jiang\footnotemark[1]}, 
 \textbf{Sendong Zhao\footnotemark[2]},
 \textbf{Bairui Hu},
 \textbf{Haochun Wang},
 \textbf{Bing Qin}
\\
\\
Harbin Institute of Technology, China
\\
  \texttt{
   \{jianboli,yjiang,sdzhao,brhu,hcwang,qinb\}@ir.hit.edu.cn
 }
}

\begin{document}
\maketitle
\footnotetext[1]{Equal Contribution.}
\footnotetext[2]{Corresponding Author.}

\renewcommand{\thefootnote}{\arabic{footnote}}

\begin{abstract}

Retrieval-Augmented Generation (RAG) helps LLMs stay accurate, but feeding long documents into a prompt makes the model slow and expensive.
This has motivated context compression, ranging from token pruning and summarization to embedding-based compression. 
While researchers have tried ``compressing'' these documents into smaller summaries or mathematical embeddings, there is a catch: the more you compress the data, the more the LLM struggles to understand it.
To address this challenge, we propose ArcAligner (\textbf{A}daptive \textbf{r}ecursive \textbf{c}ontext \textbf{Aligner}), a lightweight module integrated into the language model layers to help the model better utilize highly compressed context representations for downstream generation.
It uses an adaptive ``gating'' system that only adds extra processing power when the information is complex, keeping the system fast.
Across knowledge-intensive QA benchmarks, ArcAligner consistently beats compression baselines at comparable compression rates, especially on multi-hop and long-tail settings. 
The source code is publicly available\footnote{https://github.com/liunian-Jay/ArcAligner.git}.
\end{abstract}

\section{Introduction}


Retrieval-Augmented Generation (RAG)~\citep{lewis2020retrieval,guu2020retrieval} is now the standard approach for keeping Large Language Models (LLMs)~\citep{achiam2023gpt,touvron2023llama} factually accurate. 
By grounding answers in external documents rather than relying on memory alone, RAG helps models avoid ``hallucinations'' and answer difficult, niche questions more reliably. 
However, the standard way of doing RAG--simply pasting long documents into the prompt--creates a bottleneck. 
As we ask models to handle longer histories, deeper user profiles, or massive libraries of evidence, the input quickly becomes too long for the model to process efficiently. 
This makes effective context compression no longer just an optimization, but a necessity for building responsive, long-memory LLM agents.

\begin{figure}[t]
    \centering
    \includegraphics[width=0.995\linewidth]{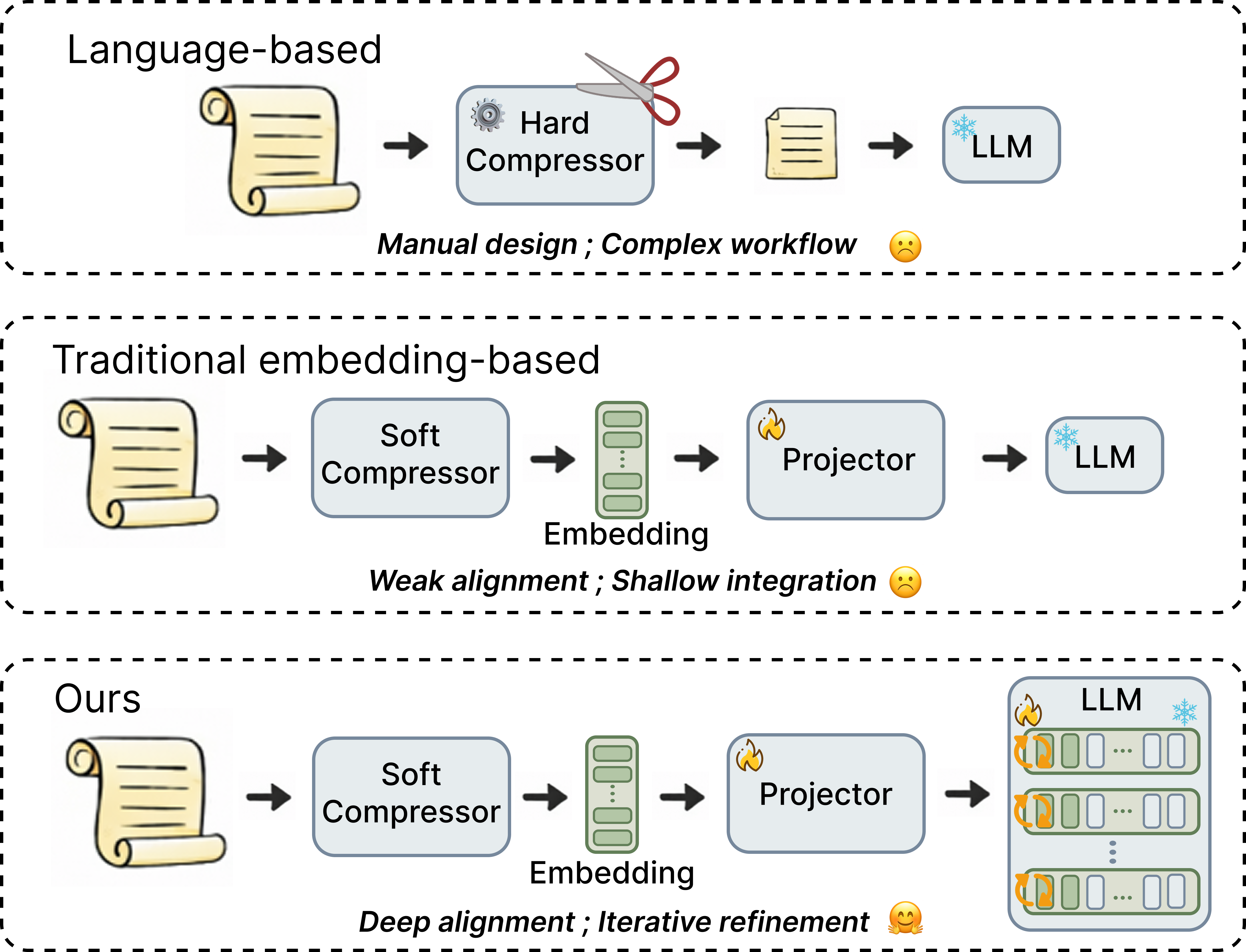}
    \caption{Comparison of different methods.}
    \label{fig:motivation}
\end{figure}

\begin{figure*}[!htp]
    \centering
    \includegraphics[width=0.995\linewidth]{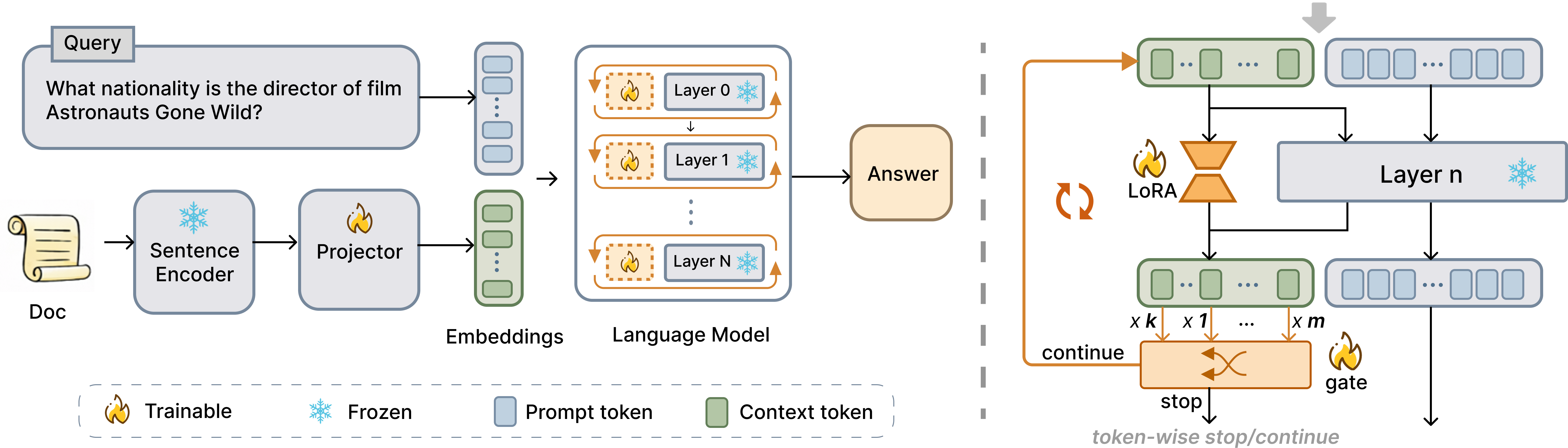}
    \caption{Illustration of the ArcAligner framework. 
    The left part is an illustration of our training and inference process, where the context is compressed and progressively refined within the LLM while the input query remains unchanged. 
    The right part is a flow of the token through each layer, passing through the LoRA and the gate, where the context tokens are adaptively and recursively refined.}
    \label{fig:framework}
\end{figure*}


To solve this problem, recent studies explore context compression--essentially shrinking the content before the LLM reads it~\citep{liu2025comprehensive}.
Current research generally follows one of two paths to achieve this. 
The first is \textbf{text-level compression}, which involves pruning unnecessary words or summarizing documents into a shorter prompt~\citep{pan2024llmlingua}. 
The second is \textbf{embedding-based compression}, which translates long text into a compact ``shorthand'' of mathematical signals (embeddings) that the model can digest quickly~\citep{cheng2024xrag,gecontext,10.1145/3701551.3703527}.
These methods suggest a compelling possibility: if we can improve an LLM’s in-model alignment to compressed context, RAG could maintain answer quality while significantly reducing time and cost.


However, as shown in Figure~\ref{fig:motivation}, these approaches still face significant limitations~\citep{nagle2024fundamental,deng2025silver,chen2025dast}.
Text-level compression is often unpredictable; because it relies on picking and choosing specific words to keep, even a small error can result in the loss of crucial evidence. 
This makes it particularly unreliable for complex, multi-step questions where every detail matters. 
On the other hand, embedding-based compression avoids deleting text but creates a ``language barrier.'' The compressed signals are mathematically different from what the LLM is used to seeing, making it difficult for the model to actually use that information to generate a grounded answer.
Despite progress on compression objectives and encoder--decoder interfaces, performance under strong compression still hinges on whether the generator can consistently exploit and progressively decode compressed evidence for answering.
We propose \textbf{ArcAligner}, a lightweight module that enables LLMs to better read and reason over compressed context representations. ArcAligner operates within the model layers to progressively align compressed signals with the representations used during generation, reducing the mismatch between compressed inputs and the model’s native processing space. It further introduces an adaptive alignment mechanism with a learned gate that controls, layer by layer, how much additional refinement is applied to each piece of compressed information. This design helps preserve task-relevant evidence and supports multi-step reasoning on complex questions. Across a range of knowledge-intensive benchmarks, ArcAligner consistently improves answer quality and reliability over prior compression-based baselines under the same context budget.
Our main contributions are:
\begin{itemize}
    \item \textbf{The ArcAligner Framework:} We propose ArcAligner, a parameter-efficient alignment framework that performs deep semantic alignment from within the language model through multi-stage training.

    \item \textbf{Adaptive Recursive Alignment:} We propose a novel ``gating'' mechanism that selectively processes information. 
    By applying additional refinement where needed, ArcAligner improves the use of compressed context without uniformly increasing computation.

    \item \textbf{Robust Performance Gains:} We provide extensive experiments and ablations showing consistent gains over strong compression baselines, 
    with clear improvements in ``difficult'' scenarios, such as multi-hop and long-tail QA.
\end{itemize}

\section{Related Work}

\subsection{Retrieval-Augmented Generation}
RAG improves factuality by conditioning LLM on retrieved evidence. 
The standard pipeline concatenates top-$k$ passages into the prompt and relies on attention to integrate evidence.
Existing approaches enhance evidence utilization with fusion-in-decoder readers~\citep{jiang2025cocoa} and end-to-end retrieval--generation training~\citep{izacard2021leveraging,izacard2023atlas,jiang2025qagent}, and also investigate reliability and controllable reliance on retrieved signals~\citep{asai2024self}.
However, long prompts incur substantial inference overhead, motivating methods that reduce the context burden while preserving evidence usability.

\subsection{Compressing Retrieved Evidence}
Context compression reduces context before it reaches the generator, via \emph{hard} compression in text space or \emph{soft} compression in embedding space.

\paragraph{Hard Compression (Text Space).}
Hard compression shortens the prompt by selecting/pruning tokens or sentences~\citep{li-etal-2023-compressing,jiang2023llmlingua,pan2024llmlingua} or rewriting/distilling passages into shorter textual evidence~\citep{xu2024recomp}. 
These methods preserve a text-only interface but require query-time decisions and can be sensitive to downstream alignment; under aggressive budgets, small mistakes may remove answer-critical evidence and disproportionately hurt difficult or long-tail questions~\citep{pan2024llmlingua}.

\paragraph{Soft Compression (Embedding Space).}
Soft compression replaces retrieved text with compact continuous representations, e.g., encoding passages into a small set of projected embedding tokens/slots~\citep{cheng2024xrag}, or injecting learned context/memory representations via reconstruction-style objectives and downstream fine-tuning~\citep{gecontext,10.1145/3701551.3703527,pilchen2025arc}. 
While avoiding long prompts, it shifts the bottleneck to alignment and usability: compressed embeddings are heterogeneous to the generator’s hidden space and typically require a projector, while training must ensure reliable decoding under strong compression~\citep{10.1145/3701551.3703527,pilchen2025arc}.

\subsection{From Compression to Usable Evidence}
A key issue is that \emph{compressing or preserving information} does not guarantee \emph{effective use} during decoding.
Related work tackles adjacent aspects: parameter-efficient adaptation (adapters, prefix/prompt tuning, low-rank updates) improves behavior under constrained or non-standard inputs~\citep{houlsby2019parameter,li2021prefix,hu2022LoRA}, while RAG-specific designs regulate reliance on retrieved evidence (e.g., self-checking/controlled use and reducing over-reliance via preference alignment)~\citep{asai2024self,jiang-etal-2025-gainrag}.
However, strong compression still poses a distinct challenge---whether compact retrieval representations can be consistently interpreted and exploited throughout decoding.
Unlike embedding-based compression methods that mainly optimize representations or compression objectives~\citep{cheng2024xrag,gecontext,10.1145/3701551.3703527,pilchen2025arc}, ArcAligner targets decode-time usability, strengthening how the generator integrates compressed retrieval information for grounded generation.

\section{Methodology}
\label{sec:method}
We propose \textbf{ArcAligner}, a parameter-efficient alignment framework (Figure~\ref{fig:framework})
that improves the usability of compressed embeddings by:
(i) enforcing \emph{layer-wise mandatory alignment} of context slots, and
(ii) enabling \emph{adaptive recursive alignment} whose depth is controlled by a learned \emph{gate}.
A three-stage training strategy progressively equips the model with reconstruction ability, task competence, and adaptive refinement.


\subsection{Problem Setup}
\label{subsec:setup}

Given a query $q$, a retriever returns the top-$k$ passages
$\{p_i\}_{i=1}^k$ from a corpus $\mathcal{P}$.
Instead of concatenating retrieved passages as text,
we compress them into a short sequence of dense
\emph{compressed context embeddings}
\[
E \in \mathbb{R}^{m \times d_r},
\]
where $m$ is the number of context slots, $d_r$ denotes the embedding dimension
of each compressed context embedding, and $|p_i|$ denotes the number of tokens
in passage $p_i$, with $m \ll \sum_i |p_i|$.
These embeddings are provided to the language model as a fixed set of designated \emph{context slots}, forming a compact interface between retrieved evidence and the base model.

\subsection{Context Slot Interface}
\label{subsec:interface}

Let the hidden size of the LLM be $d$.
We map compressed context embeddings into the model space
using a lightweight projector $W_\psi(\cdot)$, obtaining the
\emph{projected context-slot embeddings}
\begin{equation}
\tilde{E} = W_\psi(E) \in \mathbb{R}^{m \times d},
\end{equation}
where $m$ is the number of context slots and
$\tilde{E}_j \in \mathbb{R}^{d}$ denotes the embedding assigned to the $j$-th slot.

We construct an input sequence of length $n$ and reserve a fixed set of
context-slot positions
$r=\{r_1,\dots,r_m\}\subset\{1,\dots,n\}$.
Let $H^{(0,0)}\in\mathbb{R}^{n\times d}$ denote the initial token-wise hidden states
fed into the first transformer layer, defined as
\begin{equation}
H^{(0,0)}_i=
\begin{cases}
\tilde{E}_j, & \text{if } i=r_j,\\
\mathrm{Emb}(u_i), & \text{otherwise},
\end{cases}
\end{equation}
where $\mathrm{Emb}(\cdot)$ is the standard token embedding function
and $\{u_i\}_{i=1}^n$ denotes the textual input sequence
(e.g., the query and prompt tokens).

\subsection{Selective LoRA on Context Slots}
\label{subsec:selective_LoRA}

Inspired by multimodal models~\citep{dong2024internlm}, we treat context as a special modality. 
Specifically, we design context slots and apply LoRA (~\citep{hu2022LoRA}) updates only to context tokens while keeping the prompt tokens on the frozen base path. 

Let $H^{(\ell,t)}\in\mathbb{R}^{n\times d}$ denote the hidden states at
layer $\ell$ and recursive step $t$.
At the block level, let $F_\theta^{(\ell)}(\cdot)$ denote the $\ell$-th
transformer layer with frozen base parameters $\theta$,
and let $\Delta F_\phi^{(\ell)}(\cdot)$ be the LoRA-induced residual.

We define a binary broadcast mask
$M_r\in\{0,1\}^{n\times 1}$, where $(M_r)_i=\mathbb{I}[i\in r]$.
The resulting \emph{selective-LoRA} layer is defined as
\begin{equation}
\mathcal{A}^{(\ell)}(H)
=
F_\theta^{(\ell)}(H)
+
M_r \odot \Delta F_\phi^{(\ell)}(H),
\label{eq:adapter_block}
\end{equation}
where $\odot$ denotes element-wise multiplication broadcast along the hidden dimension.

\subsection{Adaptive Recursive Alignment with Gate}
\label{subsec:ara}

We introduce an \emph{Adaptive Recursive Aligner} that dynamically controls
the refinement depth of context slots across transformer layers.
A subset of layers is equipped with slot-wise gates,
denoted by $\mathcal{L}_{\text{gate}}$,
while all other layers perform a single refinement step.
Let $N$ denote the total number of transformer layers.

\paragraph{Mandatory refinement.}
At every layer $\ell\in\{0,\dots,N-1\}$, an initial refinement step is always applied:
\begin{equation}
H^{(\ell+1,0)} = \mathcal{A}^{(\ell)}\!\left(H^{(\ell,0)}\right).
\label{eq:mandatory_once}
\end{equation}

\paragraph{Gate-controlled recursive refinement.}
At gated layers ($\ell\in\mathcal{L}_{\text{gate}}$),
we allow up to $T$ additional refinement steps indexed by $t=1,\dots,T$.
Let $H_r^{(\ell+1,t-1)}\in\mathbb{R}^{m\times d}$ denote the hidden states
restricted to context-slot positions at step $t-1$.
A lightweight gating network predicts a slot-wise refinement decision:
\begin{equation}
g^{(\ell,t)}
=
\sigma\!\Big(
\mathrm{MLP}^{(\ell)}\!\big(H_r^{(\ell+1,t-1)}\big)
\Big),
\label{eq:gate_prob}
\end{equation}
where $g^{(\ell,t)}\in[0,1]^{m\times 1}$ and $\sigma(\cdot)$ denotes the sigmoid function.

Given the current state, we compute a candidate refinement by reapplying
the same selective-LoRA:
\begin{equation}
\widetilde{H}^{(\ell+1,t)}
=
\mathcal{A}^{(\ell)}\!\big(H^{(\ell+1,t-1)}\big).
\label{eq:candidate_refine}
\end{equation}

We define $\operatorname{stopgrad}(\cdot)$ as the stop-gradient operator,
which blocks gradient propagation through its argument.
During training, we adopt a straight-through estimator (STE)~\citep{bengio2013estimating} to enable
discrete refinement decisions while preserving gradient flow.
Specifically, the hard gate
$g_{\text{hard}}^{(\ell,t)}=\mathbb{I}[g^{(\ell,t)}\ge 0.5]$
is used in the forward pass, while gradients are backpropagated through
the continuous probability:
\begin{equation}
g_{\text{STE}}^{(\ell,t)}
=
g^{(\ell,t)}
+
\operatorname{stopgrad}\!\left(
g_{\text{hard}}^{(\ell,t)} - g^{(\ell,t)}
\right).
\label{eq:ste}
\end{equation}

The context-slot update is then given by
\begin{equation}
\begin{aligned}
H_r^{(\ell+1,t)}
&=
H_r^{(\ell+1,t-1)} \\
&\quad
+ g_{\text{STE}}^{(\ell,t)} \odot
\Big(
\widetilde{H}_r^{(\ell+1,t)} - H_r^{(\ell+1,t-1)}
\Big),
\end{aligned}
\label{eq:slot_update}
\end{equation}
while non-slot positions remain fixed across recursive steps:
\begin{equation}
H_{\bar{r}}^{(\ell+1,t)} = H_{\bar{r}}^{(\ell+1,0)}.
\label{eq:non_slot_fixed}
\end{equation}

At inference time, we replace $g_{\text{STE}}^{(\ell,t)}$ with the strictly
binary gate
$g_{\text{infer}}^{(\ell,t)}=g_{\text{hard}}^{(\ell,t)}\in\{0,1\}^{m\times 1}$,
so that only slots with $g_{\text{infer}}^{(\ell,t)}=1$ are further refined,
yielding an exact conditional update.

\subsection{Three-Stage Training Strategy}
\label{subsec:training}

We adopt a three-stage training strategy that progressively equips the model with
(i) alignment to compressed embeddings,
(ii) task grounding under compressed-slot inputs, and
(iii) adaptive refinement via the gate.

\paragraph{Stage I: Reconstruction Alignment Pretraining.}
We first warm up the model to interpret compressed embeddings by reconstructing the original text.
Specifically, we optimize the projector parameters $\psi$ in $W_\psi(\cdot)$ and the selective LoRA parameters $\phi$, while keeping the base LLM frozen. 
Given an input where the designated slot positions $R$ are filled with compressed embeddings $\tilde{E}$,
the model is trained to autoregressively reconstruct the context $y$
using the negative log-likelihood (NLL) objective:
\begin{equation}
\mathcal{L}_{\text{rec}}
=
-\sum_{i=1}^{|y|}
\log p_\theta\!\left(y_i \mid H^{(0,0)}, y_{<i}\right).
\label{eq:loss_rec}
\end{equation}

\paragraph{Stage II: Task-Grounded RAG Finetuning.}
Starting from Stage~I, we finetune the model on downstream RAG tasks so that it can answer queries
by grounding on compressed slot inputs.
During this stage, we continue updating $\psi$ and $\phi$ but disable the gate,
so that each layer performs only the mandatory single-step refinement.
The training target is the ground-truth answer and is optimized using the same NLL objective as in Stage~I.

\newcommand{\na}[1]{\textit{#1}} 
\newcommand{\best}[1]{\textbf{#1}}
\newcommand{\second}[1]{\underline{#1}}
\begin{table*}[!htbp]
\centering
\small
\setlength{\tabcolsep}{3.6pt}
\renewcommand{\arraystretch}{1.12}
\begin{tabular}{l c ccc ccc ccc}
\toprule
\textbf{Method} & \textbf{Comp.} &
\multicolumn{3}{c}{\textbf{2WikiMultiHopQA}} &
\multicolumn{3}{c}{\textbf{HotpotQA}} &
\multicolumn{3}{c}{\textbf{NaturalQA}} \\
& \textbf{rate} &
EM & F1 & Acc & EM & F1 & Acc & EM & F1 & Acc \\
\midrule

\rowcolor{gray!12}
\multicolumn{11}{c}{\textbf{w/o retrieval}} \\
Naive$^\nabla$ & -- &
\na{22.20} & \na{14.63} & \na{19.40} &
\na{21.20} & \na{13.39} & \na{23.20} &
\na{27.20} & \na{16.50} & \na{28.25} \\
\addlinespace[2pt]

\rowcolor{gray!12}
\multicolumn{11}{c}{\textbf{w/ retrieval}} \\
StandardRAG$^\triangle$  & -- &
\na{31.60} & \na{18.55} & \na{22.60} &
\na{37.40} & \na{19.72} & \na{36.60} &
\na{43.49} & \na{23.56} & \na{44.57} \\

xRAG & $\times$128 &
\best{48.80} & 8.66 & 22.00 &
\second{30.40} & 6.00 & 27.80 &
\best{44.13} & 3.39 & 34.04 \\

COCOM  & $\times$4 &
26.80 & 29.38 & 27.20 &
25.20 & \second{32.45} & 31.18 &
36.59 & \second{39.54} & \second{41.50} \\

COCOM  & $\times$16 &
27.80 & \second{31.02} & \second{28.80} &
24.40 & 31.77 & \second{31.20} &
35.18 & \best{41.04} & \best{41.94} \\

LLMLingua-2 & $\times$3 &
\second{33.40} & 15.78 & 23.20 &
\best{31.40} & 16.41 & 29.40 &
\second{37.73} & 18.98 & 36.95 \\

ArcAligner (Ours) & $\times$24 &
31.80 & \best{36.00} & \best{33.40} &
26.60 & \best{34.92} & \best{33.40} &
31.72 & 37.82 & 37.04 \\
\bottomrule
\end{tabular}
\caption{EM/F1/Acc on 2WikiMultiHopQA, HotpotQA, and NaturalQA.
Comp.\ rate denotes the context compression ratio. 
Best and second-best scores are highlighted in \textbf{bold} and \underline{underlined}, respectively. 
\na{Italic} denote excluded reference settings: Naive$^\nabla$ (no-retrieval) and StandardRAG$^\triangle$ (full-context).
}
\label{tab:main_results_1}
\end{table*}

\begin{table*}[t]
\centering
\small
\setlength{\tabcolsep}{3.6pt}
\renewcommand{\arraystretch}{1.12}
\begin{tabular}{l c ccc ccc ccc}
\toprule
\textbf{Method} & \textbf{Comp.} &
\multicolumn{3}{c}{\textbf{PopQA\_longtail}} &
\multicolumn{3}{c}{\textbf{TriviaQA*}} &
\multicolumn{3}{c}{\textbf{WebQuestions}} \\
& \textbf{rate} &
EM & F1 & Acc & EM & F1 & Acc & EM & F1 & Acc \\
\midrule

\rowcolor{gray!12}
\multicolumn{11}{c}{\textbf{w/o retrieval}} \\
Naive$^\nabla$ & -- &
\na{20.01} & \na{8.90} & \na{13.30} &
\na{56.60} & \na{33.25} & \na{44.80} &
\na{38.34} & \na{25.32} & \na{45.96} \\
\addlinespace[2pt]

\rowcolor{gray!12}
\multicolumn{11}{c}{\textbf{w/ retrieval}} \\
StandardRAG$^\triangle$ & -- &
\na{45.96} & \na{23.10} & \na{33.95} &
\na{69.40} & \na{35.63} & \na{60.20} &
\na{39.67} & \na{23.96} & \na{45.23} \\

xRAG & $\times$128 &
\second{33.95} & 4.94 & 28.95 &
\best{69.80} & 11.53 & 55.80 &
\best{55.86} & 5.14 & \second{47.79} \\

COCOM & $\times$4 &
24.59 & 25.15 & 24.37 &
60.80 & \second{63.60} & \second{58.80} &
\second{42.72} & 36.43 & 47.44 \\

COCOM & $\times$16 &
24.16 & \second{25.79} & 24.59 &
58.00 & 61.91 & 58.40 &
37.84 & \second{40.45} & \best{48.77} \\

LLMLingua-2 & $\times$3 &
\best{40.17} & 15.66 & \best{31.09} &
\second{66.60} & 30.62 & 51.80 &
37.75 & 21.04 & 42.77 \\

ArcAligner (Ours) & $\times$24 &
30.38 & \best{32.80} & \second{30.74} &
62.80 & \best{65.65} & \best{62.80} &
35.14 & \best{43.19} & 46.80 \\
\bottomrule
\end{tabular}
\caption{EM/F1/Acc on PopQA\_longtail, TriviaQA*, and WebQuestions.
Comp.\ rate denotes the context compression ratio. 
Best and second-best scores are highlighted in \textbf{bold} and \underline{underlined}, respectively. 
\na{Italic} denote excluded reference settings: Naive$^\nabla$ (no-retrieval) and StandardRAG$^\triangle$ (full-context).
}
\label{tab:main_results_2}
\end{table*}

\paragraph{Stage III: Gate-Aware Adaptive Finetuning.}
In the final stage, we enable the gating mechanism and train the full adaptive recursive aligner. 
We jointly optimize the projector parameters $\psi$, the selective LoRA parameters $\phi$, and the gate parameters.
The forward pass follows \S~\ref{subsec:ara}, using STE-based gating during training and hard binary gating at inference. 
The optimization in this stage still uses NLL loss. Through training in this stage, ArcAligner is trained more thoroughly and can learn slot refinement depth according to task requirements, thereby supporting more effective inference under strong compression.

\section{Experiments}

\subsection{Implementation Details}
\label{subsec:implementation}

\paragraph{Base Models.}
We use \emph{Mistral-7B-Instruct-v0.2}~\citep{jiang2023mistral7b} as the base language model in all main experiments. 
We use \emph{SFR-Embedding-Mistral}~\citep{meng2024sfrembedding} as the sentence encoder to encode passages into dense vectors.
All parameters of the base model are frozen during training, except for the lightweight modules introduced by our method. 

\paragraph{Passage Segmentation.}
We segment retrieved passages at the sentence level using spaCy~\citep{honnibal2020spacy}.
These sentence-based segments are used as the basic units for compression. 

\begin{table*}[!t]
\centering
\small
\setlength{\tabcolsep}{4.5pt}
\renewcommand{\arraystretch}{1.10}
\begin{tabular}{l ccc ccc ccc}
\toprule
Method &
\multicolumn{3}{c}{HotpotQA} &
\multicolumn{3}{c}{NaturalQA} &
\multicolumn{3}{c}{TriviaQA*} \\
& EM & F1 & Acc & EM & F1 & Acc & EM & F1 & Acc \\
\midrule
ArcAligner (Ours)
& \second{26.60} & \best{34.92} & \best{33.40}
& \best{31.72} & \best{37.82} & \best{37.04}
& 62.80 & \second{65.65} & \best{62.80} \\
\ \ \ w/o Recursion
& 26.00 & \second{34.48} & \second{32.60}
& 31.39 & \second{37.64} & \second{36.90}
& 62.00 & 65.17 & 62.00 \\
\ \ \ w/o Gate (Max Loop)
& \best{27.20} & 31.69 & 31.00
& 31.25 & 32.57 & 32.74
& \textbf{64.00} & 62.66 & 61.80 \\
\ \ \ w/o LoRA \& Recursion
& 25.00 & 32.02 & 32.00
& 31.58 & 37.50 & 36.76
& \second{63.40} & \best{66.20} & \second{62.20} \\
\bottomrule
\end{tabular}
\caption{Ablation results on HotpotQA, NaturalQA, and TriviaQA*. We report non-strict EM, F1, and
LLM-judged accuracy (Acc). \textbf{w/o Recursion} disables gate-controlled refinement loops at inference.
\textbf{w/o Gate (Max Loop)} disables the gate while keeping a fixed number of refinement
loops for all context slots. \textbf{w/o LoRA \& Recursion} disables the selective-adaptation weights
(LoRA) and refinement loops, but keeps the context slots.}

\label{tab:ablation}
\end{table*}

\paragraph{Training Data.}
Our training follows a three-stage training strategy. 
\textit{\textbf{Stage I (Pretraining).}}
Pretraining data is built from a large-scale Wikipedia corpus. We follow standard preprocessing by segmenting the dump into fixed-length passages, then sample about 200k passages and train for one epoch. This stage learns a robust alignment between embeddings and the language model’s representation space.
\textit{\textbf{Stage II \& III (RAG Finetuning).}}
Both stages are trained on the HotpotQA~\citep{yang2018hotpotqa} dataset, using approximately 90K training samples.
Stage II enables only the projector and LoRA parameters, while Stage III further enables the gate for adaptive refinement. Details can be found in Appendix~\ref{sec:appendix-details}.




\subsection{Datasets and Evaluation Metrics}
\label{subsec:datasets_metrics}



\paragraph{Evaluation Datasets.} 
We evaluate effectiveness and generalization on multi-hop and open-domain QA benchmarks. For multi-hop QA, we use 2WikiMultiHopQA~\citep{ho2020constructing} and HotpotQA~\citep{yang2018hotpotqa}; for open-domain QA, we use NaturalQA~\citep{kwiatkowski2019natural}, PopQA\_longtail~\citep{mallen2023not}, TriviaQA*~\citep{joshi2017triviaqa}, and WebQuestions~\citep{berant2013semantic}, where \emph{TriviaQA*} denotes a sub-sampled evaluation set consisting of 500 randomly selected questions.
For each dataset, we retrieve the top-20 passages with \emph{Contriever}~\citep{izacard2021unsupervised} and rerank them with \emph{RankZephyr}~\citep{pradeep2023rankzephyr}. 
The reranked top passage is used as the context. 
All datasets share the same retrieval and inference settings, with details in Appendix~\ref{sec:appendix-datasets}.


\paragraph{Evaluation Metrics.}
We report Exact Match (\textbf{EM}), \textbf{F1}, and Accuracy (\textbf{Acc}). Following prior work~\citep{asai2024self,mallen2023not}, we use \emph{non-strict} EM, counting a prediction as correct if it contains the gold answer. F1 is computed as token-level overlap with the gold answer, complementing EM since longer outputs can increase EM but lower F1. We additionally report LLM-judged accuracy; details are in Appendix~\ref{sec:appendix-evaluation}.


\subsection{Baselines}
We compare against representative baselines: (1)~\textbf{Naive}, which answers the question without any retrieval context; (2)~\textbf{StandardRAG}, the classic retrieve-then-read setup that concatenates retrieved passages with the query; (3) \textbf{xRAG}~\citep{cheng2024xrag}, which replaces text passages with projected dense retrieval embeddings for extreme compression;(4) \textbf{COCOM}~\citep{10.1145/3701551.3703527}, which injects learned context embeddings into the decoder; and (5) \textbf{LLMLingua-2}~\citep{pan2024llmlingua}, which compresses retrieved contexts by retaining only informative tokens before generation. 

\subsection{Main Results}
\label{subsec:main_results}

Table~\ref{tab:main_results_1} and~\ref{tab:main_results_2} report the main results. Some key findings are as follows.

\textbf{(1) Compression vs. No Compression. }
StandardRAG achieves the highest EM and accuracy on most datasets, setting a performance upper bound by providing the full textual context. In contrast, embedding-based compression methods (e.g., xRAG and COCOM) show a notable drop, highlighting the limitations of current compression techniques. 
Our approach offers a trade-off and performs well on multi-hop and long-tail tasks.

\textbf{(2) Language space vs. Embedding space.} 
Overall, language-based methods showed mixed results, suggesting they require unique and complex designs to handle different tasks. Embedding-based methods outperformed LLMlingua-2 overall and demonstrated stronger robustness across various tasks. 
Our ArcAligner exhibits even greater robustness and learn more task-oriented information.

\textbf{(3) Shallow Alignment vs. Deep alignment}
Shallow alignment methods such as xRAG exhibit high EM but low F1, resulting in excessively long outputs, poor task orientation, and even worse performance. In contrast, deep alignment methods like COCOM and ArcAligner not only ensure information availability but also provide greater task orientation.

\subsection{Ablation Study}
\label{sec:ablation}
We conduct ablations on two core components of ArcAligner, with results reported in Table~\ref{tab:ablation}.

Disabling refinement loops (\textit{w/o Recursion}) consistently degrades performance, indicating that recursive refinement provides larger gains than a single pass. 
Further removing the gating mechanism while keeping a fixed number of refinement loops for all context slots (\textit{w/o Gate (Max Loop)}) leads to more pronounced drops on NaturalQA and TriviaQA*, suggesting that indiscriminate refinement is suboptimal and that the gate is crucial for selecting which context slots to refine.

Finally, removing selective adaptive weights while retaining projection-generated context slots (\textit{w/o LoRA \& Recursion}) performs worse than the full model, highlighting the importance of deep semantic alignment.

Overall, layer-wise selective adaptation and learned gated recursion are key to aligning compressed context embeddings.

\subsection{Analysis Across Different Backbones}
\label{sec:backbone_analysis}

\begin{figure}[t]
    \centering
    \includegraphics[width=0.95\linewidth]{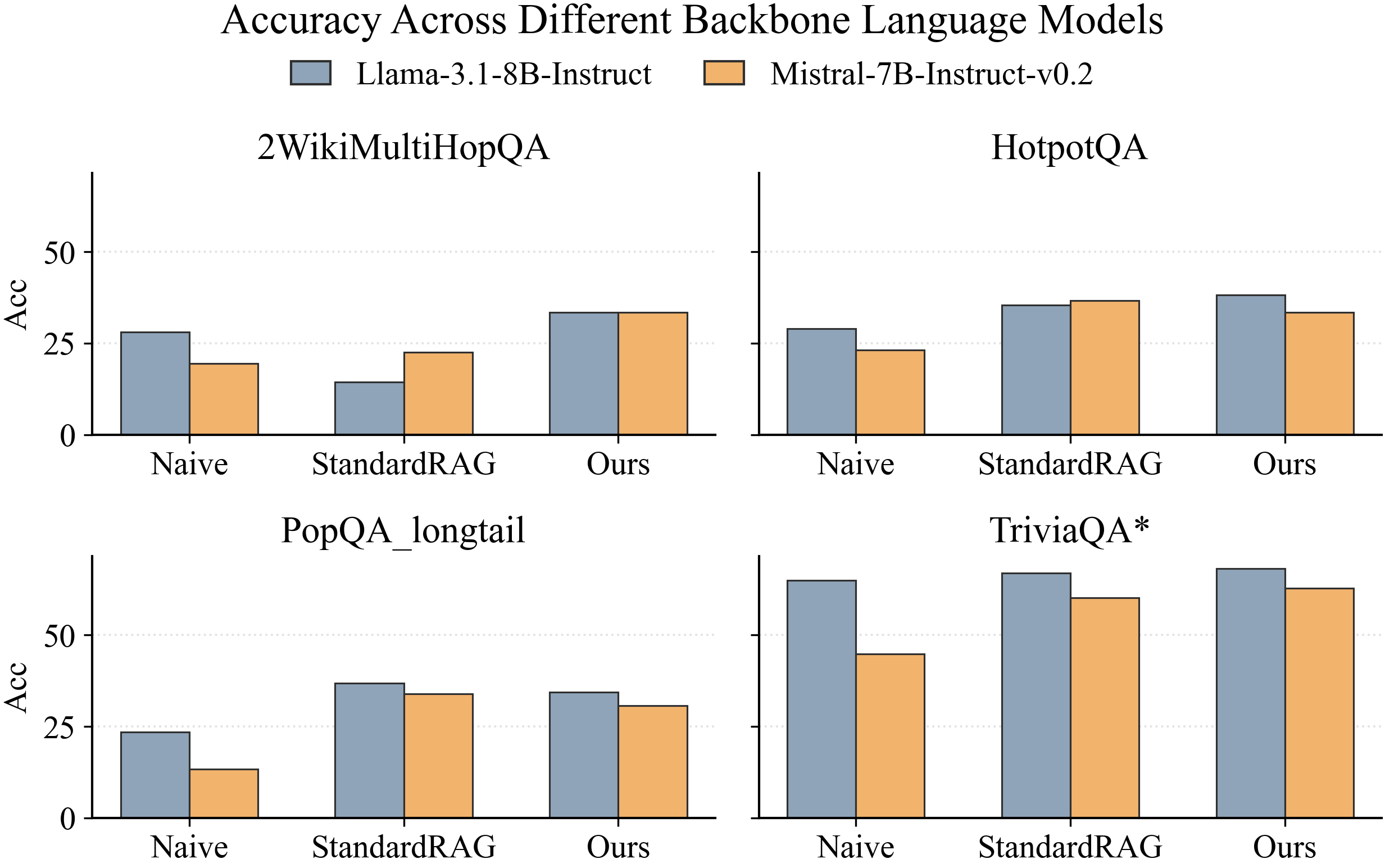}
    \caption{
    Accuracy across different backbone language models on four QA datasets.
    Grouped bars compare \textbf{Naive}, \textbf{StandardRAG}, and \textbf{ArcAligner(Ours)} under two backbones:
    \textbf{Llama-3.1-8B-Instruct} and \textbf{Mistral-7B-Instruct-v0.2}.
    }
    \label{fig:backbone_accuracy}
\end{figure}

Figure~\ref{fig:backbone_accuracy} reports the accuracy of different methods under two backbone language models on four representative QA datasets.
Across datasets, the relative performance ordering among \textit{Naive}, \textit{StandardRAG}, and ArcAligner remains largely consistent when switching between Llama-3.1 and Mistral-v0.2, although the absolute accuracy varies across backbones.
In particular, \textit{ArcAligner} exhibits comparable behavior under both backbone models and maintains competitive performance across datasets.
These results suggest that the observed method-level trends are not specific to a particular backbone language model.

\subsection{Reconstruction Ability After Pretraining}
\label{sec:reconstruction_ppl}

To investigate reconstruction pre-training, we compare the reconstruction perplexity (PPL) on the test set after the pre-training stage.

\begin{figure}[t]
    \centering
    \includegraphics[width=\linewidth]{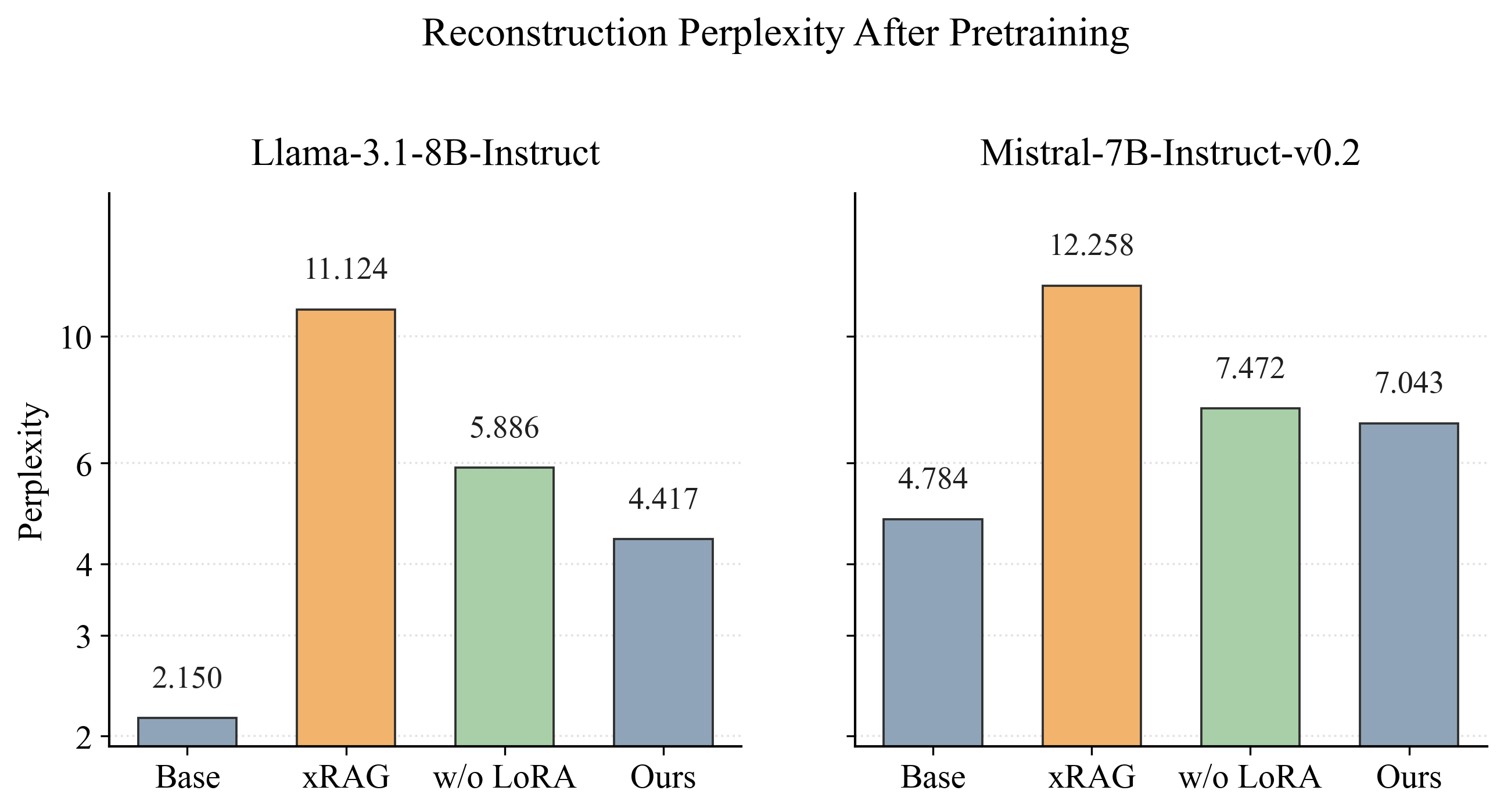}
    \caption{
    Reconstruction perplexity (PPL) after the pretraining stage, evaluated on a held-out pretraining test set with 10k samples.
    Results are reported for two backbone language models, \textbf{Llama-3.1-8B-Instruct} and \textbf{Mistral-7B-Instruct-v0.2}.
    Lower PPL indicates better reconstruction fidelity.
    }
    \label{fig:reconstruction_ppl}
\end{figure}

Figure~\ref{fig:reconstruction_ppl} reports the results. \textit{Base} represents the model directly reciting the context without compression, and can be considered as the upper limit of reconstruction performance. 
\textit{xRAG} yields substantially higher PPL for both backbone language models, indicating that directly applying retrieval-based compression markedly reduces reconstruction fidelity during pre-training. 
Our model attains lower PPL than \textit{xRAG} and \textit{w/o LoRA}, showing that the proposed alignment design mitigates the degradation introduced by compression. 
Removing the deep alignment module (\textit{w/o LoRA}) also lowers PPL relative to \textit{xRAG}, but remains significantly worse than \textit{Base}. This suggests that shallow projection alone is insufficient for the LLM to interpret compressed embeddings.

\begin{figure}[!b]
    \centering
    \includegraphics[width=0.95\linewidth]{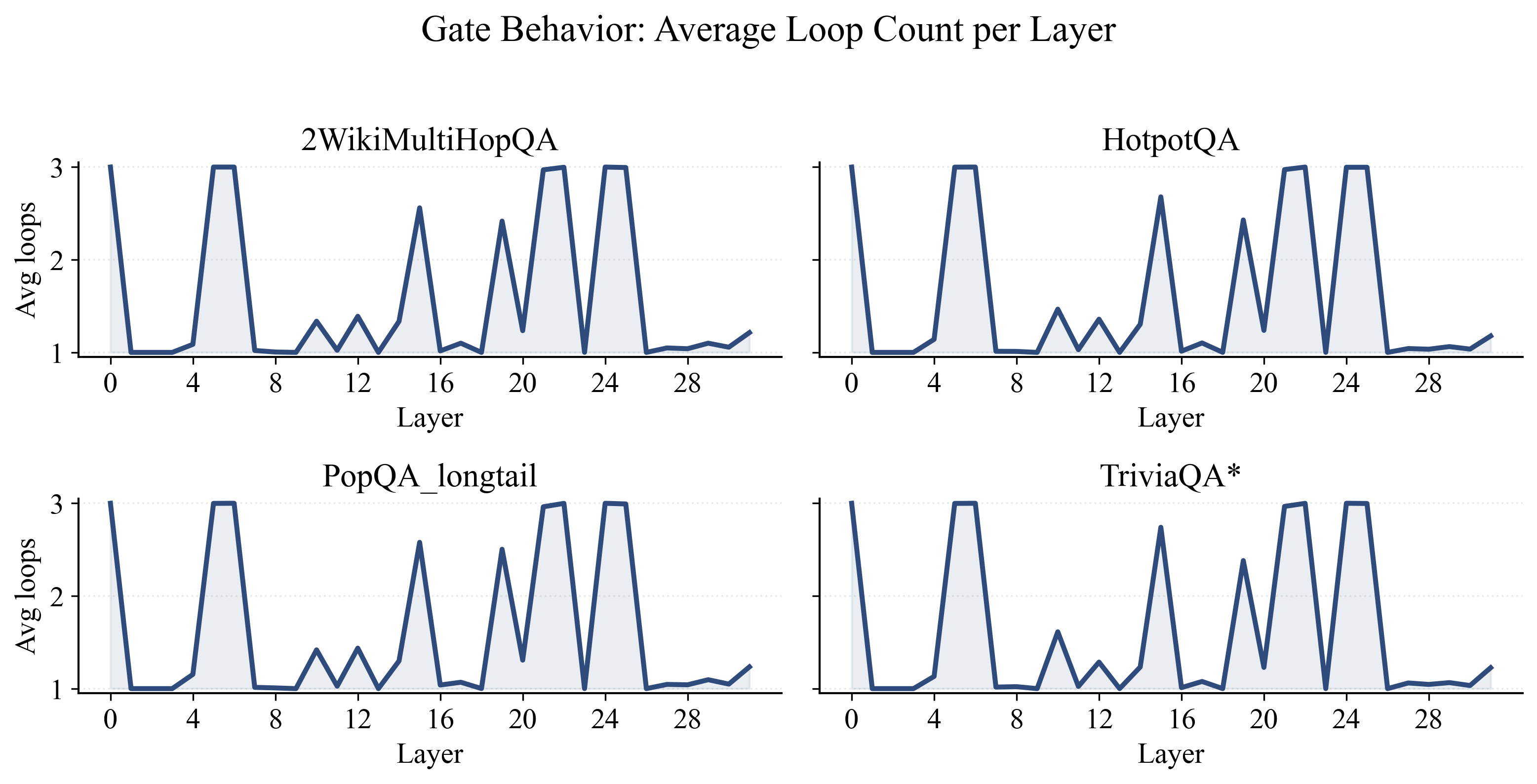}
    \caption{
    Gate behavior analysis across layers.
    The figure shows the token-level average number of refinement loops triggered by the gate at each layer evaluated on four datasets.}
    \label{fig:gate_behavior}
\end{figure}
\subsection{Analysis of Gate Behavior}
\label{sec:gate_behavior}
To investigate the layer-by-layer behavior of gates, we compute the average number of refinement loops triggered per layer.

As shown in Figure~\ref{fig:gate_behavior}, the gates exhibit a consistent pattern across all four datasets: most layers operate with a single forward pass, while a few layers selectively activate additional refinement loops. This suggests that the gate allocates extra computation at specific stages of the network, rather than uniformly increasing depth across layers. Notably, higher loop counts are concentrated in the early and late layers, whereas the middle layers remain relatively stable, approaching a single forward pass. We hypothesize that early layers are crucial for embedding alignment, while later layers support semantic fusion, which benefits QA. More interpretable mechanisms remain an important direction for further study.

\subsection{4-way Error Category Analysis}
\label{sec:4way-analysis}

\begin{figure}[!b]
    \centering
    \includegraphics[width=0.95\linewidth]{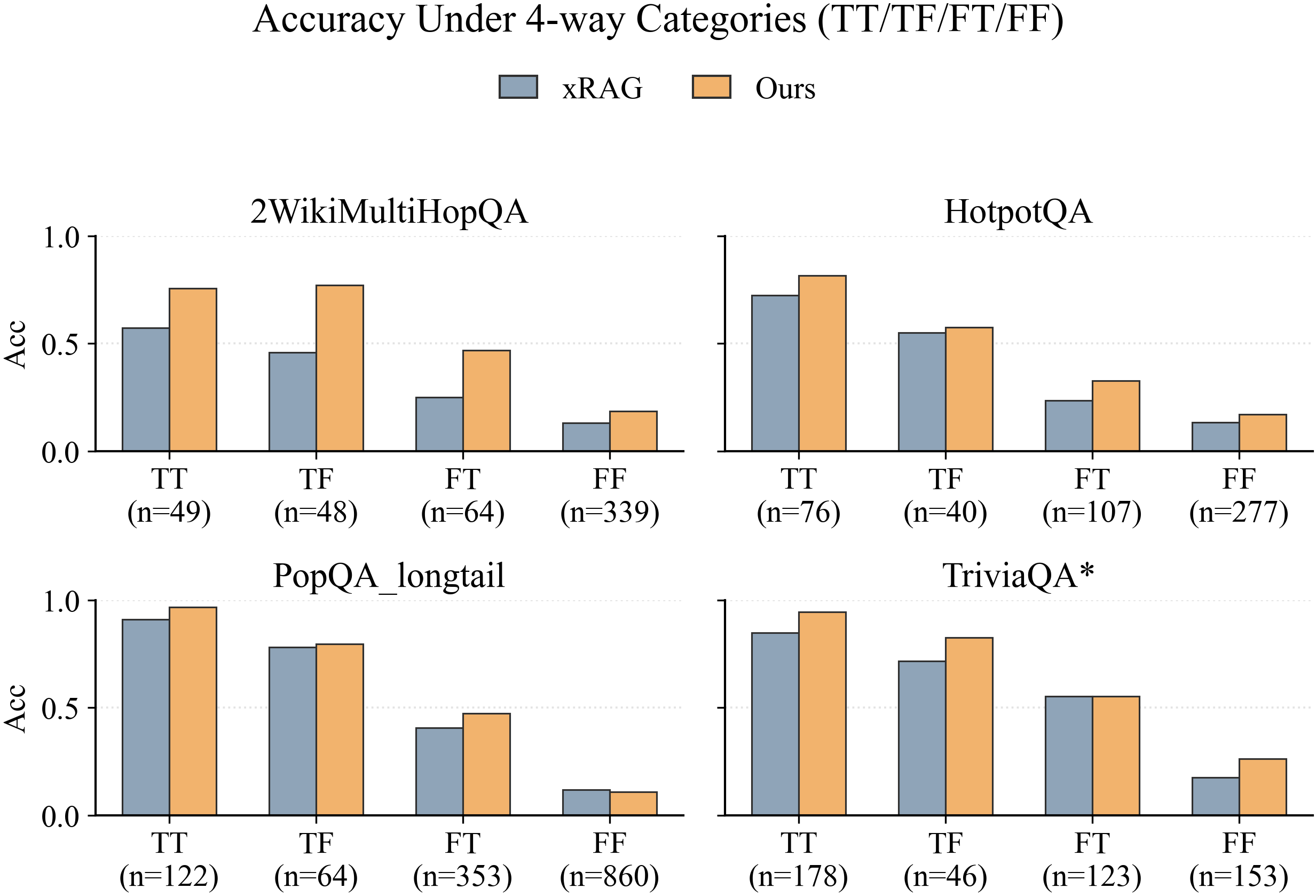}
    \caption{\textbf{4-way category analysis} on four datasets. We partition questions into four groups based on whether \textit{naive prompting} and \textit{standard RAG} answer correctly: \textbf{TT}, \textbf{TF}, \textbf{FT}, and \textbf{FF}, where \textbf{T/F} denote correct/incorrect predictions. We report accuracy (\textit{Acc}) of xRAG and ArcAligner within each category. Category size is shown under each label (e.g., \textit{TT} with $n{=}41$).}
    \label{fig:4way-category}
\end{figure}

To better understand where ArcAligner improves, we partition questions into four categories based on whether the \textit{Naive} and \textit{StandardRAG} succeed: \textbf{TT}, \textbf{TF}, \textbf{FT}, and \textbf{FF}. The results are shown in Figure~\ref{fig:4way-category}. 
We observe several consistent trends.

First, in the \textbf{TT} category, where both the naive prompt and StandardRAG succeed, ArcAligner achieves strong performance across all datasets. This indicates that ArcAligner does not introduce harmful behavior on simpler or well-supported questions.  In the \textbf{FT} category, ArcAligner consistently outperforms \emph{xRAG}, suggesting that its compression preserves useful information better than comparable approaches. 

Notably, ArcAligner shows clear advantages in the \textbf{TF} category, which typically corresponds to cases where retrieved passages are insufficient or contain misleading signals. Compared with \emph{xRAG}, ArcAligner produces a much higher proportion of correct responses, indicating that it can better compensate for imperfect retrieval rather than merely relying on retrieved content. 
The \textbf{FF} category is the most challenging. ArcAligner still yields a small number of correct cases, which we attribute to the later training stages that encourage the LLM to attend more effectively to informative signals in the compressed context.

Overall, benefiting from adaptive recursion and three-stage training, ArcAligner improves performance and exhibits robustness for RAG tasks.

\subsection{Effect of Compression Ratio}
\label{sec:compression_ratio}

\begin{figure}[t]
    \centering
    \includegraphics[width=0.95\linewidth]{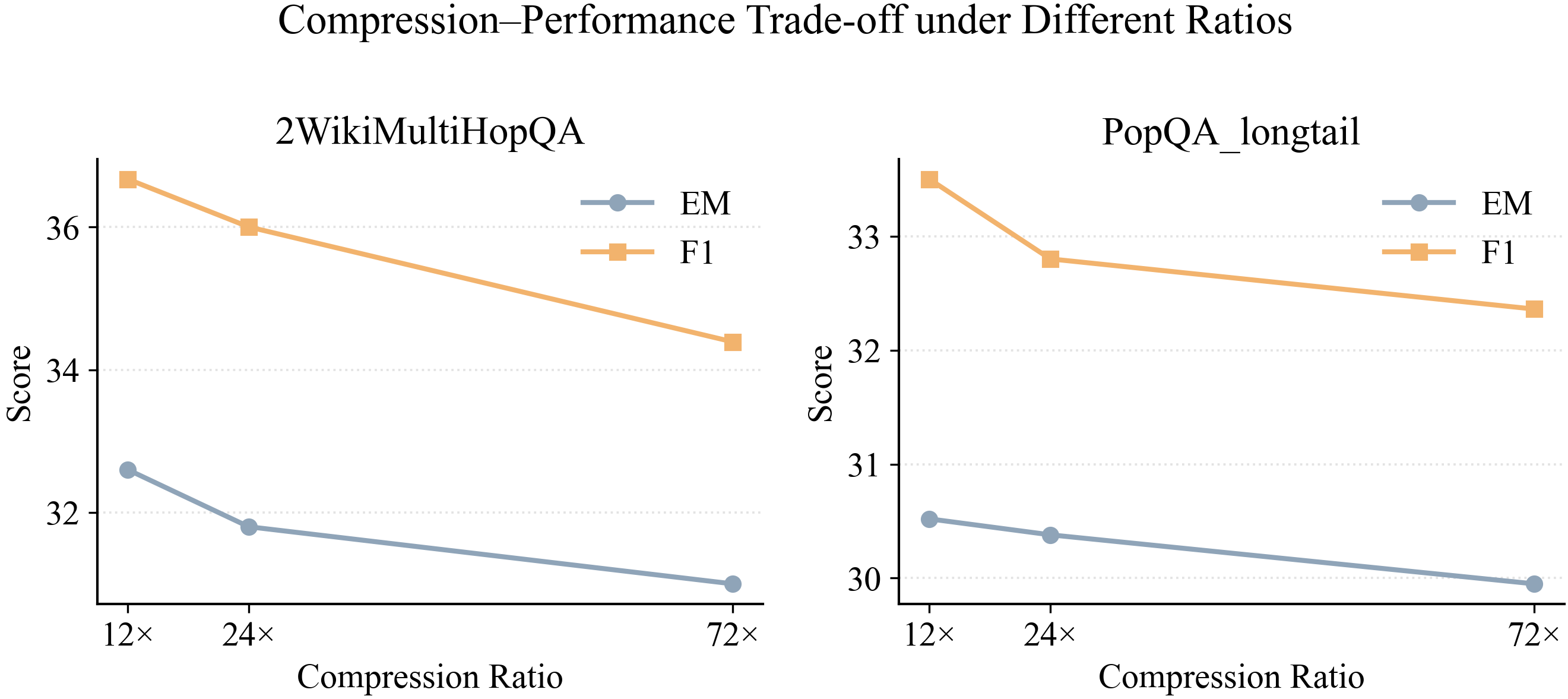}
    \caption{
    Performance under different compression ratios on 2WikiMultiHopQA and PopQA\_longtail. EM/F1 scores are used as evaluation metrics. 
    }
    \label{fig:compression_ratio}
\end{figure}

We further analyse the impact of different compression ratios on ArcAligner. For each ratio, we train a separate model with the same training settings, and vary only the sentence segmentation granularity that controls the ratio: 12$\times$ is obtained by splitting each sentence into two segments, whereas 72$\times$ is obtained by merging three consecutive sentences into one, with all other settings fixed.

As shown in Figure~\ref{fig:compression_ratio}, at a higher compression ratio of 72x, performance decreased slightly, but not significantly. This indicates that the compressed representation retains the key evidence required for multi-hop inference.
Meanwhile, we found some performance improvement as we reduce the compression ratio, suggesting that a smaller compression ratio can retain more key information and mitigate excessive information loss. 
This suggests that in practice, we need to flexibly adjust the compression ratio to achieve a trade-off between performance and efficiency.

\section{Conclusion}
\label{sec:conclusion}
In this paper, we propose ArcAligner, a parameter-efficient framework that enhances the usability of compressed context embeddings for RAG. By introducing an adaptive recursive alignment mechanism with gating, ArcAligner ensures deep semantic alignment between compressed retrieval signals and the language model’s internal representations. 
Extensive experiments on knowledge-intensive QA benchmarks show that ArcAligner consistently outperforms existing compression baselines, especially in multi-hop and long-tail QA tasks, making it a valuable tool for efficient RAG systems.

\section*{Limitations}
We discuss the limitations of ArcAligner as follows. Our current setup assumes a fixed context slot interface and a specific compression pipeline, and we do not systematically study how design choices such as passage segmentation, compression ratio, or slot budget affect the accuracy–efficiency trade-off. We also focus on a constrained retrieval setting and do not fully explore Top-$K$ multi-document evidence, where complementary information across documents can be crucial; how to allocate slots and refinement budget across multiple retrieved passages remains open. Finally, ArcAligner relies on hard, slot-wise gating with a maximum recursion depth $T$ at selected layers; while efficient in typical cases, the gate placement and $T$ are tuned heuristically and may be suboptimal under strict compute/latency constraints, motivating future work on budget-aware gating or adaptive stopping.

\bibliography{custom}

\appendix

\section{Datasets}
\label{sec:appendix-datasets}
Here, we introduce in detail the datasets we used, which are seven datasets on four tasks.

\textbf{2WikiMultiHopQA}~\citep{ho2020constructing} and \textbf{HotpotQA}~\citep{yang2018hotpotqa}:
Both are multi-hop question answering benchmarks constructed from Wikipedia articles.
To control experimental cost, we follow prior work~\citep{trivedi2023interleaving,kim2024sure}
and evaluate on the released sub-sampled splits, each containing 500 questions drawn
from the original validation set.

\textbf{NaturalQA}~\citep{kwiatkowski2019natural}:
A large-scale benchmark designed to support comprehensive question answering.
The questions are collected from real Google search queries, and the answers are
annotated as text spans from corresponding Wikipedia articles by human annotators.

\textbf{PopQA-longtail}~\citep{mallen2023not}:
An open-domain QA dataset targeting long-tail factual knowledge.
The questions are entity-centric and automatically generated from Wikidata knowledge
triples using relation-specific templates.
Entity popularity is quantified via Wikipedia page views, enabling controlled evaluation
on low-popularity (long-tail) facts.

\textbf{TriviaQA*}~\citep{joshi2017triviaqa}:
A collection of open-domain trivia questions paired with answer annotations,
originally sourced from online trivia websites.
We use \emph{TriviaQA*} to denote a sub-sampled evaluation set of 500 randomly selected questions.

\textbf{WebQuestions}~\citep{berant2013semantic}:
A factoid question answering dataset constructed from real user queries issued to
the Google Suggest API, where answers correspond to specific entities in Freebase.

Dataset statistics are summarized in Table~\ref{tab:datasets}

\begin{table}[!h]
\centering
\resizebox{0.85\linewidth}{!}{%
\begin{tabular}{@{}ccc@{}}
\toprule
Task Type             & Datasets      & \# Samples \\ 
\hline
\multirow{2}{*}{Multi-HopQA} & 2WikiMultiHopQA & 500    \\
                             & HotpotQA        & 500    \\
\midrule
\multirow{3}{*}{OpenQA}      & NaturalQA       & 3610   \\
                             & PopQA\_longtail & 1399   \\
                             & TriviaQA*       & 500    \\
                             & WebQuestions    & 2032   \\
\bottomrule
\end{tabular}
}
\caption{Description of tasks and evaluation datasets.}
\label{tab:datasets}
\end{table}

\section{Evaluation}
\label{sec:appendix-evaluation}

\paragraph{Evaluation Metrics.}
We report three metrics: Exact Match (EM), F1, and Accuracy (Acc).

For EM and F1, we follow the standard evaluation protocol used in prior RAG work.
Exact Match measures whether the model’s prediction contains the gold answer.
Following Self-RAG~\cite{asai2024self} and When Not to Trust LMs~\cite{mallen2023not},
we adopt a \emph{non-strict} EM metric, where a prediction is considered correct if it includes
the gold answer, rather than requiring an exact string match.
F1 measures the token-level overlap between the predicted answer and the gold answer.

As noted in prior work, longer responses may artificially improve EM due to higher matching
probability, while often reducing F1 due to the inclusion of irrelevant content.
Therefore, EM and F1 provide complementary perspectives on answer quality.

In addition, we report Accuracy (Acc) based on large language model judgment.
Specifically, we use \textbf{Qwen3-30B-A3B-Instruct-2507}~\citep{yang2025qwen3} as an evaluator to assess whether the
generated answer correctly addresses the question given the reference answer.
The evaluator is prompted to produce a binary correctness judgment, which is then averaged
over the dataset to compute Acc.

\section{Implementation Details}
\label{sec:appendix-details}

\subsection{Hyperparameters}
For all experiments, we adopt instruction-tuned variants of the base language models.
All training and evaluation are conducted on a cluster of \textbf{4 NVIDIA H20-3e GPUs}.

Unless otherwise specified, all models are trained using the AdamW optimizer with linear
learning rate scheduling. We report the detailed hyperparameter configurations for different training stages in Tables~\ref{tab:pretrain_hparams} and~\ref{tab:finetune_hparams}.

\begin{table}[t]
\centering
\begin{tabular}{l c}
\toprule
\textbf{Hyperparameter} & \textbf{Assignment} \\
\midrule
optimizer & AdamW \\
learning rate & 2.0e-4 \\
lr scheduler type & linear \\
warmup ratio & 0.03 \\
weight decay & 0.0 \\
LoRA r & 128 \\
LoRA alpha & 32 \\
LoRA dropout & 0.05 \\
max loops & 3 \\
loop layers & all \\
epochs & 1 \\
batch size & 8 \\
gradient accumulation steps & 8 \\
num GPUs & 4 \\
max train samples & 200,000 \\
\bottomrule
\end{tabular}
\caption{Hyperparameters for Stage I.}
\label{tab:pretrain_hparams}
\end{table}

\begin{table}[t]
\centering
\begin{tabular}{l c}
\toprule
\textbf{Hyperparameter} & \textbf{Assignment} \\
\midrule
optimizer & AdamW \\
learning rate & 2.0e-5 \\
lr scheduler type & linear \\
warmup ratio & 0.03 \\
weight decay & 0.0 \\
LoRA r & 128 \\
LoRA alpha & 32 \\
LoRA dropout & 0.05 \\
max loops & 3 \\
loop layers & all \\
epochs & 1 \\
batch size & 8 \\
gradient accumulation steps & 2 \\
num GPUs & 4 \\
max train samples & 90,447 \\
\bottomrule
\end{tabular}
\caption{Hyperparameters for Stage II \& III.}
\label{tab:finetune_hparams}
\end{table}

\subsection{Prompts for Stage I}
\label{app:paraphrase_pretrain}

The list of instructions for paraphrase pretraining is shown in Table~\ref{tab:paraphrase_pretrain_instr}.
They present the same meaning with natural language variance.
Following prior work on embedding-based context compression, these paraphrase-style instructions
are adapted from the instruction templates used in xRAG~\citep{cheng2024xrag}.

\begin{table*}[t]
\centering
\small
\setlength{\tabcolsep}{10pt}
\renewcommand{\arraystretch}{1.15}
\begin{tabularx}{\textwidth}{@{}p{0.22\textwidth}X@{}}
\toprule
\textbf{Instruction intent} & \textbf{Template (with placeholders)} \\
\midrule

Equivalence statement &
\texttt{Background: \textcolor{phgreen}{[B]}.}
\ \texttt{This is equivalent to: \textcolor{phorange}{[T]}.} \\[2pt]

Direct rewrite &
\texttt{Rewrite the background in your own words: \textcolor{phgreen}{[B]} $\rightarrow$ \textcolor{phorange}{[T]}.} \\[2pt]

Restatement request &
\texttt{Provide a restatement of the background: \textcolor{phgreen}{[B]}.}
\ \texttt{Return: \textcolor{phorange}{[T]}.} \\[2pt]

Paraphrase-as-question &
\texttt{\textcolor{phgreen}{[B]} is a paraphrase of what?}
\ \texttt{Answer with: \textcolor{phorange}{[T]}.} \\[2pt]

Two-formulation alignment &
\texttt{These two expressions convey the same meaning:}
\ \texttt{(1) \textcolor{phgreen}{[B]} \quad (2) \textcolor{phorange}{[T]}.} \\[2pt]

Minimal-output constraint &
\texttt{Restate \textcolor{phgreen}{[B]} using a single sentence.}
\ \texttt{Output only \textcolor{phorange}{[T]}.} \\

\bottomrule
\end{tabularx}

\vspace{2pt}
\caption{Stage-I instruction templates.
\textcolor{phgreen}{[B]} and \textcolor{phorange}{[T]} denote the compressed background representation and the target textual reconstruction, respectively.
We sample from multiple intent categories to diversify supervision while keeping a consistent I/O format.}
\label{tab:paraphrase_pretrain_instr}
\end{table*}

\subsection{Prompt for Stage II \& III}
\label{app:finetune_prompt}

During task fine-tuning, we represent the compressed background document using a short instruction line composed of repeated background placeholders.
This line is inserted at a designated location in the prompt and marks the positions of the injected compressed segments:

\begin{quote}
\textit{Refer to the background document:} \textcolor{phgreen}{[B]} \textcolor{phgreen}{[B]} \dots \textcolor{phgreen}{[B]} \\
\textit{Question:} \textcolor{phorange}{[Q]}
\end{quote}

The number of background placeholders is set to \(N=\max(1,|\texttt{sentences}|)\), and the instruction line is instantiated with \(N\) copies of \textcolor{phgreen}{[B]}.

\section{Algorithm}
\label{sec:appendix-algorithm}

This section provides the detailed pseudocode of the ArcAligner forward computation,
corresponding to the description in Section~\ref{sec:method}.
The algorithm specifies how gated recursive updates are applied at selected layers
and context-slot positions during both training and inference, as detailed in
Algorithm~\ref{alg:arcaligner}.

\begin{algorithm}[t]
\caption{ArcAligner Forward}
\label{alg:arcaligner}
\begin{algorithmic}[1]
\Require Initial states $H^{(0,0)}$, blocks $\{\mathcal{A}^{(\ell)}\}_{\ell=0}^{N-1}$,
gated layers $\mathcal{L}_{\text{gate}}$, max recursion $T$,
mode $\in\{\texttt{train},\texttt{infer}\}$
\Ensure Final states $H^{(N,0)}$
\Statex \textit{$H|_{r}$: context-slot positions;\quad $H|_{\bar r}$: non-slot positions}

\For{$\ell = 0$ \textbf{to} $N-1$}
  \State $H^{(\ell+1,0)} \gets \mathcal{A}^{(\ell)}(H^{(\ell,0)})$
  \If{$\ell \in \mathcal{L}_{\text{gate}}$}
    \For{$t = 1$ \textbf{to} $T$}
      \State $g \gets \sigma(\mathrm{MLP}^{(\ell)}(H^{(\ell+1,t-1)}|_{r}))$
      \If{$\texttt{mode}=\texttt{train}$}
        \State $g \gets g + \mathrm{stopgrad}(\mathbb{I}[g \ge 0.5] - g)$
      \Else
        \State $g \gets \mathbb{I}[g \ge 0.5]$
      \EndIf
      \State $\widetilde{H} \gets \mathcal{A}^{(\ell)}(H^{(\ell+1,t-1)})$
      \State $H^{(\ell+1,t)}|_{r} \gets H^{(\ell+1,t-1)}|_{r}
      + g \odot (\widetilde{H}|_{r} - H^{(\ell+1,t-1)}|_{r})$
      \State $H^{(\ell+1,t)}|_{\bar r} \gets H^{(\ell+1,0)}|_{\bar r}$
    \EndFor
    \State $H^{(\ell+1,0)} \gets H^{(\ell+1,T)}$
  \EndIf
\EndFor
\State \Return $H^{(N,0)}$
\end{algorithmic}
\end{algorithm}

\section{Case Studies}
\label{sec:appendix-cases}
\newcommand{\CleanCaseBox}[1]{%
  \noindent\fcolorbox{black!70}{white}{%
    \begin{minipage}{0.97\linewidth}
    \vspace{6pt}
    #1
    \vspace{6pt}
    \end{minipage}
  }%
}
We present several representative cases to further analyze the behavior of our method
(Figures~\ref{fig:case1_toronto_nw}--\ref{fig:case3_astronauts_gone_wild}).\\

\noindent\textbf{Notation.}
We use \texttt{traj} to visualize the per-slot recursion depth determined by the gate.
A token \texttt{L} corresponds to one recursion step, and the number of repeated \texttt{L}'s matches the loop count reported in \texttt{loops}.
Thus, \texttt{L0.} means the default single pass (loop count $=1$), whereas \texttt{LLL} means three recursive passes (loop count $=3$).

\begin{figure*}[t]
\centering

\CleanCaseBox{
\textbf{Question.} In what country is Toronto Northwest?

\vspace{0.4em}
\textbf{Compressed Sentence Passages.}
\begin{enumerate}\setlength{\itemsep}{2pt}\setlength{\parsep}{0pt}\setlength{\topsep}{2pt}
    \item was redistributed between Davenport, Spadina, Trinity and York West ridings.
    \item \textbf{Toronto Northwest was a federal electoral district represented in the House of Commons of Canada from 1925 to 1935.}
    \item It was located in the city of Toronto in the province of Ontario.
    \item This riding was created in 1924 from parts of Parkdale, Toronto North and York South ridings.
    \item It consisted of the part of the city of Toronto north of Bloor Street, west of Bathurst St. and east of the Northern Division of the Canadian National Railway.
\end{enumerate}

\vspace{0.4em}
\textbf{Answer.} Canada.

\vspace{0.4em}
\textbf{Routing Behavior.} At layer 31:
\[
\texttt{traj} = [\text{L0.},\ \mathbf{LLL},\ \text{L0.},\ \text{L0.},\ \text{L0.}]
\]
The answer evidence appears in the \textbf{second sentence}, which is repeatedly updated via recursive routing (\texttt{LLL}).
}

\caption{Case 1: the answer evidence appears in the second sentence, which is repeatedly updated via recursive routing.}
\label{fig:case1_toronto_nw}
\end{figure*}

\begin{figure*}[t]
\centering
\CleanCaseBox{
\textbf{Question.} What sport does Radik Zhaparov play?

\vspace{0.4em}
\textbf{Compressed Sentence Passages.}
\begin{enumerate}\setlength{\itemsep}{2pt}\setlength{\parsep}{0pt}\setlength{\topsep}{2pt}
    \item \textbf{Radik Zhaparov (born February 29, 1984) is a Kazakh ski jumper who has competed since 2003.}
    \item At the 2006 Winter Olympics in Turin, he finished 11th in the team large hill and 26th in the individual normal hill events.
    \item At the FIS Nordic World Ski Championships, Zhaparov has finished 11th in team events three times and 24th in the individual normal hill events.
    \item Zharparov's best individual World Cup finish was 11th in a large hill event in Finland in 2007.
    \item His best individual career finish was second in an FIS Cup normal.
\end{enumerate}

\vspace{0.4em}
\textbf{Answer.} ski jumping.

\vspace{0.4em}
\textbf{Routing Behavior.} At layer 31:
\[
\texttt{traj} = [\mathbf{LLL},\ \text{L0.},\ \text{L0.},\ \text{L0.},\ \text{L0.}]
\]
The recursive routing (\texttt{LLL}, loop count $=3$) is assigned to the first slot, which corresponds to the \textbf{first sentence} containing the explicit evidence.
}
\caption{Case 2: the answer evidence appears in the first sentence, which receives repeated recursive updates (\texttt{LLL}) at layer 31.}
\label{fig:case2_radik}
\end{figure*}

\begin{figure*}[t]
\centering
\CleanCaseBox{
\textbf{Question.} What nationality is the director of film \emph{Astronauts Gone Wild}?

\vspace{0.4em}
\textbf{Compressed Sentence Passages.}
\begin{enumerate}\setlength{\itemsep}{2pt}\setlength{\parsep}{0pt}\setlength{\topsep}{2pt}
    \item \textbf{\emph{Astronauts Gone Wild} is a 2004 documentary video produced and directed by Bart Sibrel, a Nashville, Tennessee-based video maker.}
    \item Sibrel made this video as a follow-up to his 2001 video \emph{A Funny Thing Happened on the Way to the Moon}.
    \item The title of the presentation is a wordplay on the \emph{Girls Gone Wild} video series.
    \item In \emph{Astronauts Gone Wild}, Sibrel confronts nine Apollo \ldots
\end{enumerate}

\vspace{0.4em}
\textbf{Answer.} American.

\vspace{0.4em}
\textbf{Routing Behavior.} At layer 14:
\[
\texttt{traj} = [\mathbf{LLL},\ \text{L0.},\ \text{L0.},\ \text{L0.}]
\]
The recursive routing (\texttt{LLL}, loop count $=3$) is assigned to the first slot, which aligns with the \textbf{first sentence} describing the director and location cues used to infer nationality.
}
\caption{Case 3: the director evidence appears in the first sentence, which receives repeated recursive updates (\texttt{LLL}) at layer 14.}
\label{fig:case3_astronauts_gone_wild}
\end{figure*}

\end{document}